\pdfoutput=1

\documentclass[11pt]{article}

\usepackage[]{acl}

\usepackage{times}
\usepackage{latexsym}
\usepackage{booktabs}
\usepackage{multirow}
\usepackage{linguex}
\usepackage{tablefootnote}
\usepackage{url}

\usepackage[font=small]{caption}

\usepackage[shortlabels]{enumitem}
\usepackage[normalem]{ulem}

\usepackage[T1]{fontenc}

\usepackage[utf8]{inputenc}

\usepackage{microtype}
\usepackage{array}
\usepackage{graphicx}
\usepackage{booktabs}
\usepackage{listings}
\usepackage{xcolor}
\usepackage[skip=5pt]{caption}

\usepackage[bordercolor=black, backgroundcolor=white, linecolor=black]{todonotes}
\title{Information Extraction from Conversation Transcripts: \\ Neuro-Symbolic vs. LLM}

\author{
  \textbf{Alice Saebom Kwak\textsuperscript{1}},
  \textbf{Maria Alexeeva\textsuperscript{2}},
  \textbf{Gus Hahn-Powell\textsuperscript{2}},
  \textbf{Keith Alcock\textsuperscript{2}},\\
  \textbf{Kevin McLaughlin}\textsuperscript{2},
  \textbf{Doug McCorkle}\textsuperscript{3},
  \textbf{Gabe McNunn}\textsuperscript{3},
  \textbf{Mihai Surdeanu\textsuperscript{2}}
\\
\\
  \textsuperscript{1} Department of Linguistics, University of Arizona\\
  \textsuperscript{2} Lum AI\\
  \textsuperscript{3} Eocene Environmental Group\\
  \small{
    \{alicekwak0520, maxaalexeeva\}@gmail.com, \{ghp, keith, kevin, mihai\}@lum.ai, \newline\{dmccorkle, gmcnunn\}@eocene.com}}
\vspace{20mm}

\begin{document}
\maketitle
\begin{abstract}

The current trend in information extraction (IE) is to rely extensively on large language models, effectively discarding decades of experience in building symbolic or statistical IE systems. This paper compares a neuro-symbolic (NS) and an LLM-based IE system in the agricultural domain, evaluating them on nine interviews across pork, dairy, and crop subdomains. The LLM-based system outperforms the NS one (F1 total: 69.4 vs. 52.7; core: 63.0 vs. 47.2), where total includes all extracted information and core focuses on essential details. However, each system has trade-offs: the NS approach offers faster runtime, greater control, and high accuracy in context-free tasks but lacks generalizability, struggles with contextual nuances, and requires significant resources to develop and maintain. The LLM-based system achieves higher performance, faster deployment, and easier maintenance but has slower runtime, limited control, model dependency and hallucination risks. Our findings highlight the ``hidden cost'' of deploying NLP systems in real-world applications, emphasizing the need to balance performance, efficiency, and control.
\end{abstract}

\section{Introduction}

Information extraction (IE) from written text is a well-established task, with various systems successfully extracting entities, relations, and events from unstructured natural language text \cite{sundheim1991overview}. Recently, IE approaches have increasingly relied on large language models (LLMs) (see \citet{xu2024large} for a comprehensive survey). While this shift has led to notable performance improvements, it comes at a cost: (i) low explainability \cite{danilevsky_2021_explainability_for_nlp}, (ii) fragility \cite{Sculley2015HiddenTD}, and (iii) lack of control over its behavior \cite{vacareanu2024softrules}. These challenges contribute to the ``hidden technical debt'' of deploying natural language processing (NLP) systems in real-world industrial settings. In contrast, rule-based methods are explainable and provide greater control, but lack the generalization power of current deep learning systems \cite{Tang2022ItTT}.

This work investigates whether the performance boost of LLMs justifies the hidden costs in the real world. We compare two radically different IE strategies---an LLM-based system and an equivalent neuro-symbolic (NS) implementation---in an industry application extracting information from agricultural conversation transcripts. These interviews, conducted between farmers (interviewees) and interviewers, gather details about U.S. farms (e.g., number of animals raised, crop types). Our use case, while highly specialized, can be generalized to other domains like customer support systems. To encourage further research in this underexplored setting, we release our dataset: \url{https://anonymous.4open.science/r/ag_dataset-8E0B/}

Our comparison highlights key trade-offs: the NS system is efficient, offers greater control, and is precise in context-free tasks but lacks generalizability, struggles with contextual nuances, and requires extensive resources for development and maintenance. The LLM-based system offers higher performance, faster deployment, and lower maintenance costs but is slower, provides less control, is model dependent, and is prone to hallucinations. We hope that this discussion will influence the design of future IE systems that are deployed and maintained in the real world.

\section{Related Works}

Information Extraction (IE) is a key task in NLP that has been investigated with various approaches, including rule-based (\citealp{hobbs-etal-1993-fastus};  \citealp{10.3115/1119089.1119095}; and \citealp{cunningham-etal-2002-gate}), neural (\citealp{socher-etal-2012-semantic}; \citealp{zeng-etal-2014-relation}, \citealp{chen-etal-2015-event}; \citealp{nguyen-grishman-2015-relation}; and \citealp{zeng-etal-2015-distant}), and LLM-based methods (\citealp{josifoski-etal-2022-genie}; \citealp{ma-etal-2023-query}; \citealp{wadhwa-etal-2023-revisiting}; \citealp{wadhwa-etal-2023-revisiting}; \citealp{perot-etal-2024-lmdx}; \citealp{hu-etal-2025-large}). Several studies also combine or compare different approaches (\citealp{gardner-etal-2013-improving}; \citealp{10.1145/2736277.2741687}; \citealp{liu-etal-2020-multi}; \citealp{imaichi-etal-2013-comparison}; and \citealp{wang2024does}; See appendix \ref{related_works} for full discussion).

Our work builds on these directions in two distinct ways. First, it focuses on a dialogue-based IE task within the agricultural domain, using interview transcripts that are long, noisy, and contextually rich. These characteristics introduce new challenges for both neuro-symbolic and LLM-based systems. Rules often fail to account for conversational variation, while LLMs can struggle with long input windows. The presence of domain-specific terminology further complicates both automatic speech recognition (ASR) and information extraction. Second, we compare neuro-symbolic and LLM-based systems in an industrial setting. While most prior work focuses on academic benchmarks, we evaluate the approaches in terms of deployment feasibility, robustness, and efficiency, addressing the gap between research and real-world application identified by \citet{chiticariu-etal-2013-rule}.

\section{The task}

\begin{figure}[htbp]
    \centering

    \includegraphics[width=2in]{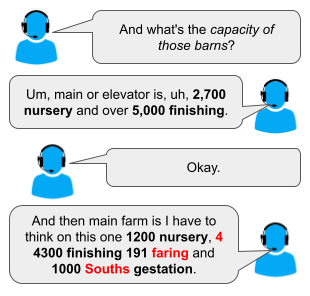}
    \caption{Sample conversation transcript simplified for readability, presented as produced by the ASR system before postprocessing. The target identifier to be extracted is in italics. The target values for the identifier are in bold. 
    The ASR errors are in red: the first \textit{4} is spurious, \textit{faring} must be corrected to \textit{farrowing}, and \textit{Souths} to \textit{sows}.'
    }
    \label{fig:canon-example}
\end{figure}

We extract structured information from semi-scripted interviews between farmers and representatives of an organization\footnote{Anonymized for review} that supports sustainability efforts by tracking farm data (e.g., farm facilities, acreage, crop types, etc., see Figure \ref{fig:canon-example}).
 Interviewers collect information through telephone or video calls loosely following a script tailored to the farm type (e.g., pork, dairy, crop). 

Collected information includes \textit{identifiers} such as `barn capacity' (see Figure \ref{fig:canon-example}), `manure storage', and `renewable energy'. Identifiers serve as variables with different value types: quantitative (e.g., number of animals per barn), categorical (e.g., lagoons vs. pits for manure storage), or boolean (e.g., use of renewable energy). The project's goal is to expedite data collection by automatically extracting variable-value pairs from interview transcripts, reducing reliance on manual note-taking.


\vspace{-7pt}

\section{Approach}



\begin{figure*}[th!]
    \centering
    \includegraphics[width=5.4in]{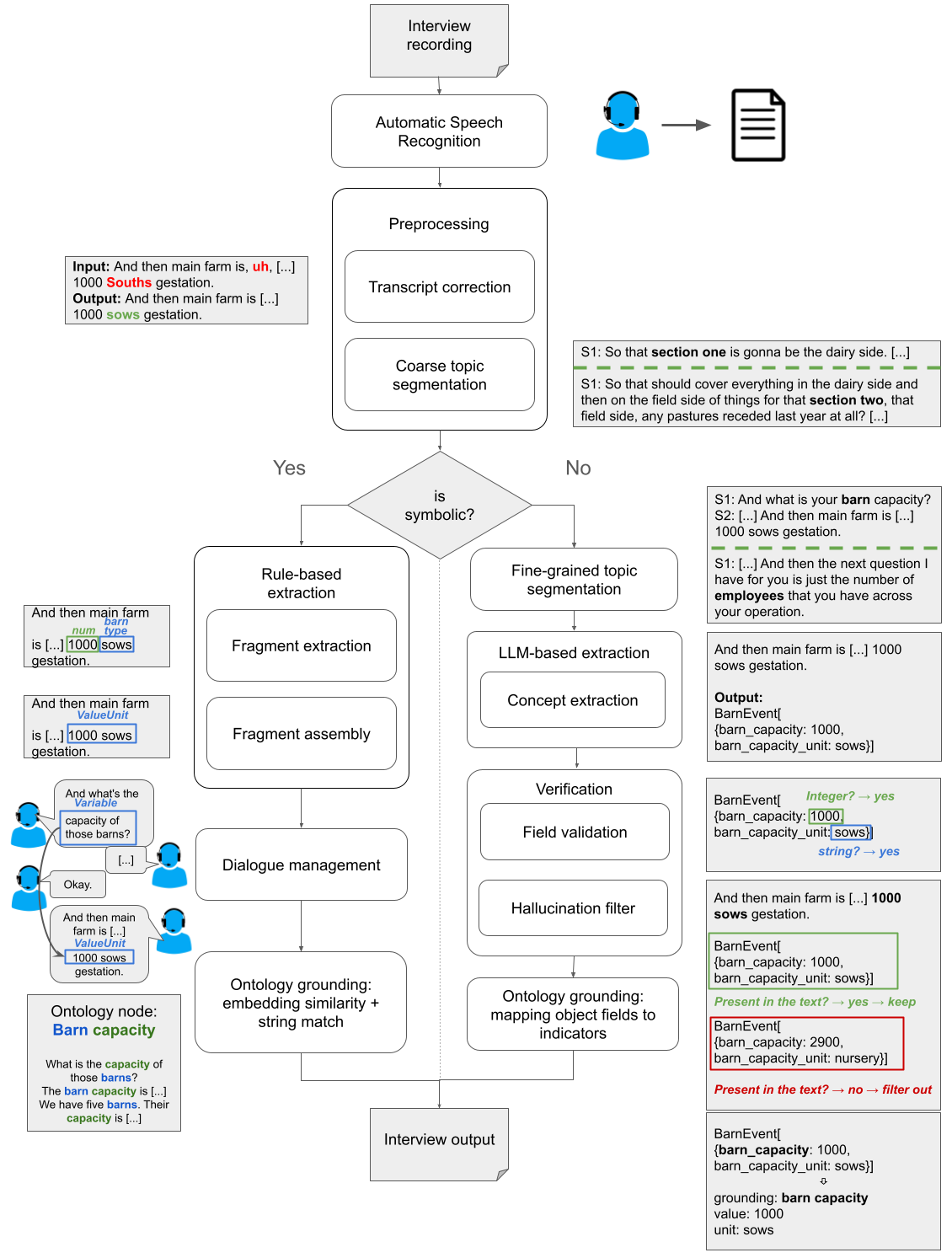}
    \caption{Generalized pipeline of the two system versions. Both share ASR and preprocessing but diverge afterward. The NS system uses a rule-based approach to extract and assemble identifier-value pairs, with dialogue management linking distant fragments. Grounding relies on embedding similarity and string matching. The LLM-based system segments interviews by topic, extracts information via in-context learning, and verifies results through field validation and hallucination filtering before mapping fields to relevant indicators.}
    \label{fig:system}
\end{figure*}

Our system (Figure \ref{fig:system}) aims to link target variables in interviewer questions to values in farm representatives' responses. To normalize extractions, values are mapped to an ontology (e.g., `barn capacity' is equivalent to `capacity of those barns' and can be linked to the same identifier). Both NS and LLM-based versions share ASR and preprocessing but diverge in extraction and ontology grounding.

\vspace{-5pt}



\subsection{The shared components}



\subsubsection{Automatic speech recognition (ASR)}

Interviews are conducted and recorded via a video-conferencing platform. We convert the audio recordings to text using an ASR service provided by rev.ai\footnote{\url{https://rev.ai}}, chosen for its favorable user agreement, including data privacy protections. The ASR service converts the audio to text, providing timestamps and marking speaker turns. Additionally, some atmospheric tags (e.g., <laugh>, <affirmative>) are embedded within the text.

\subsubsection{Preprocessing}

\paragraph{Transcript correction:} Since the ASR model is not domain-specific, its output requires post-processing. This includes remapping frequently mistranscribed domain-specific terms (e.g., \textit{South} to \textit{sows} in Figure \ref{fig:canon-example}) and normalizing spelling in unambiguous cases (e.g., adding hyphens in \textit{farrow to finish} and \textit{18 46 0}, which refers to the fertilizer \textit{18-46-0}). This normalization helps our NS extraction system. We also remove filler words (e.g., \textit{um}).



\paragraph{Topic segmentation: } In the agriculture domain, we work with several subdomains (dairy, crops, and pork), each associated with an identifier ontology that may overlap. To improve accuracy, we segment interviews using interviewer section markers (e.g., ``\textbf{Section one} is about dairy''), which indicate topic transitions. The system then assigns each segment to its relevant domain.



\subsection{Neuro-symbolic system}

The neuro-symbolic system starts with an encoder-based backbone. In particular, we implemented a multi-task learning framework, in which an encoder is shared between multiple low-level NLP tasks: part-of-speech tagging, named-entity recognition, and syntactic dependency parsing. Each of these tasks is implemented as a sequence model using a dedicated linear classifier head. To convert dependency parsing into sequence modeling, we used the algorithm of \citet{vacareanu-etal-2020-parsing}, which converts each dependency tree into two sequences of: (a) relative positions of the governor tokens for each token in the sentence, and (b) labels of the dependencies between the each token and its governor.

\label{symbolic}
\subsubsection{Rule-based extraction} 

For symbolic information extraction, we develop a grammar using the Odin information extraction framework \cite{valenzuela-escarcega-etal-2015-domain} to write rules over semantic and syntactic features. We first extract standalone entities: identifiers and values (\textit{Fragment extraction} in Figure \ref{fig:system}). A sample rule can be found in Appendix \ref{entityrule}. Extracting identifiers and categorical values relies on domain-specific knowledge bases, which require updates for new subdomains.

    

We assemble extracted entities into identifier-value relations (\textit{Fragment assembly} in Figure \ref{fig:system}). Identifiers and their values are frequently scattered across multiple sentences, requiring complex dialogue management to properly assemble.


\subsubsection{Dialogue management}

Identifier and values may span multiple speaker turns, with identifiers appearing in questions and values distributed across responses (Figure \ref{fig:canon-example}). To address this, we search prior context for a matching identifier when extracting a potential value (e.g., \textit{1200 nursery} → \textit{capacity of those barns}).


To improve precision, we use heuristics such as adjusting context window size per identifier and maintaining valid identifier-unit pairings (e.g., \textit{nursery} pairs with \textit{barn capacity} but not \textit{manure storage}). A fallback system assigns unambiguous values when no identifier is available. (e.g., compound values that include barn types such as \textit{10 farrowing} can be unambiguously assigned to \textit{barn capacity}).






\subsubsection{Ontology grounding}

For ontology grounding, we map each identifier-value pair to an
ontology using a combination of vector embedding similarity and text string overlap. This way, the identifiers in Figure \ref{fig:system} (e.g., \textit{capacity of those barns}, \textit{barn capacity}) can be linked to the same ontology node---`barn capacity.'


\subsection{LLM approach}
\label{llmbased}


\subsubsection{Fine-grained topic segmentation}
\label{sec:fine_segmentation}

In the LLM-based approach, we segment interviews into smaller blocks based on topic shifts. For example, if the discussion transitions from barn capacity to manure management, we split the interview accordingly. This segmentation reduces token usage and minimizes hallucinations by narrowing the expected information types for each block.

Topic changes are detected using predefined keywords or key phrases. For instance, if `barn' appears in a speaker's turn, the topic likely shifts to barn capacity or type. A curated keyword list guides this segmentation process. We use a keyword-based approach to give users greater control and facilitate easier maintenance. While statistical block segmentation methods may perform better, we chose this strategy as it gives users the possibility to customize the keyword list as needed to align with their specific requirements.

\begin{table*}[h]
\centering
\small
\footnotesize
\renewcommand{\arraystretch}{0.78}
\begin{tabular}{lcccccc}
\toprule
\textbf{Domain} & \multicolumn{2}{c}{\textbf{Precision}} & \multicolumn{2}{c}{\textbf{Recall}} & \multicolumn{2}{c}{\textbf{F1}} \\
               & \textbf{Total} & \textbf{Core} & \textbf{Total} & \textbf{Core} & \textbf{Total} & \textbf{Core} \\
\midrule
\multicolumn{7}{l}{\textbf{Neuro-symbolic System}} \\
\midrule
Pork  & 68.5 {\scriptsize (58.1–76.9)} & 62.4 {\scriptsize (51.9–70.0)} & 54.6 {\scriptsize (46.2–69.0)} & 47.9 {\scriptsize (40.0–60.9)} & 60.6 {\scriptsize (51.4–72.7)} & 54.0 {\scriptsize (45.2–65.1)} \\
Crop  & 63.8 {\scriptsize (50.0–87.8)} & 61.3 {\scriptsize (50.0–82.1)} & 42.8 {\scriptsize (36.0–46.8)} & 38.8 {\scriptsize (36.0–43.8)} & 50.7 {\scriptsize (41.9–61.0)} & 46.6 {\scriptsize (41.9–50.5)} \\
Dairy & 67.6 {\scriptsize (50.0–85.7)} & \textbf{63.2} {\scriptsize (50.0–80.0)} & 36.4 {\scriptsize (33.3–41.9)} & 31.2 {\scriptsize (26.3–34.1)} & 46.8 {\scriptsize (40.0–51.7)} & 41.0 {\scriptsize (39.6–43.4)} \\
\midrule
\textbf{Average} & 66.6 & 62.3 & 44.6 & 39.3 & 52.7 & 47.2 \\
\midrule
\multicolumn{7}{l}{\textbf{LLM-based System}} \\
\midrule
Pork  & \textbf{73.9} {\scriptsize (71.1–77.1)} & \textbf{69.5} {\scriptsize (64.5–75.8)} & \textbf{65.1} {\scriptsize (57.4–75.0)} & \textbf{61.4} {\scriptsize (55.6–71.8)} & \textbf{68.9} {\scriptsize (65.9–74.2)} & \textbf{64.9} {\scriptsize (59.7–70.0)} \\
Crop  & \textbf{73.3} {\scriptsize (61.9–83.3)} & \textbf{65.1} {\scriptsize (54.3–77.8)} & \textbf{65.7} {\scriptsize (62.5–69.7)} & \textbf{58.5} {\scriptsize (56.0–60.0)} & \textbf{69.0} {\scriptsize (63.4–72.1)} & \textbf{61.1} {\scriptsize (56.7–65.1)} \\
Dairy & \textbf{70.4} {\scriptsize (58.8–82.3)} & 62.7 {\scriptsize (53.5–77.4)} & \textbf{71.4} {\scriptsize (57.6–87.5)} & \textbf{65.1} {\scriptsize (52.9–80.0)} & \textbf{70.4} {\scriptsize (58.2–77.8)} & \textbf{63.0} {\scriptsize (53.2–69.1)} \\
\midrule
\textbf{Average} & \textbf{72.5} & \textbf{65.8} & \textbf{67.4} & \textbf{61.7} & \textbf{69.4} & \textbf{63.0} \\
\bottomrule
\end{tabular}
\caption{Performance comparison of neuro-symbolic and LLM-based systems across domains, reporting total (all extracted information) and core (essential information) scores. LLM outperforms NS system, especially in F1 (69.4 vs. 52.7 total, 63.0 vs. 47.2 core) and recall. Scores include 95\% confidence intervals from bootstrap resampling; bold indicates the higher score.}
\label{tab:totalres}
\end{table*}

\subsubsection{LLM-based extraction}

For LLM-based extraction, we use in-context learning with task instructions and expected output formats. Each target information type has a corresponding Pydantic dataclass~\citep{Colvin_Pydantic_2025}, which defines detailed extraction guidelines, including content specifications and format requirements. We formulate prompts that incorporate the JSON schema of each target class, guiding the model toward structured generation with constrained decoding to produce valid JSON that matches the dataclass structure. (See Appendices \ref{sec:prompt} and \ref{sec:dataclass} for the full prompt details and a sample dataclass).

We use OpenChat 3.5, employing a quantized version for efficiency\footnote{We used openchat-3.5-0106.Q5\_K\_M.gguf, available at: \url{https://huggingface.co/TheBloke/openchat-3.5-0106-GGUF}}. The model was selected after a pilot study comparing quantized versions of LLaMA-3, Mixtral 8x7B, and OpenChat 3.5. Despite being the smallest, OpenChat 3.5 outperformed the others, likely due to its dialogue-specific tuning, making it well-suited for our task. We only considered local models, as proprietary options present privacy risks. The model has a maximum token limit of 8,192, and the temperature was set to 0 to ensure deterministic outputs.

\subsubsection{Verification}

Once the LLM generates raw output, it undergoes verification. Extracted fields are validated, and if a field has an invalid type, an attempt is made to convert it (e.g., `true' is converted to True for boolean fields). If conversion is not possible, the value is filtered out. Additionally, to prevent hallucination, extracted strings are checked against the input text. If no overlap is found, the value is considered a hallucination and is discarded.

\subsubsection{Ontology grounding}

After post-processing, all validated fields are mapped onto the ontology. This mapping is guided by a CSV file that links fields to corresponding indicators. For example, the field `barn\_capacity' in the BarnEvent dataclass is mapped to barn capacity in the ontology as specified in the CSV file. Once mapping is complete, the output is transformed into its final format, which includes the grounding, grounding ID, value, and optional unit.

\vspace{-6pt}

\section{Experiments}

We compare two versions of the system: an NS version (Section \ref{symbolic})
and an LLM-based version (Section \ref{llmbased}).
In this section, we describe the experiment settings, the results, and the error analysis. 

\vspace{-6pt}

\subsection{Setting}

We evaluate the two systems using nine interviews, three per subdomain. While the dataset comprises nine interviews, each is lengthy and information-dense---averaging 28 minutes 27 seconds, 4,700 words\footnote{See Appendix \ref{word-count-statistics} for word count statistics per interview.}, and 189 speaker turns---yielding a richly structured, domain-specific corpus. Gold data was manually created, and system outputs were assessed using exact match, reporting precision, recall, and F1 scores.

For a fair comparison, we report two scores: total and core. Total counts all extracted information, treating variants like \textit{wheat} and \textit{winter wheat} as separate true positives without penalizing systems for extracting only one. Core uses stricter criteria, grouping such variants as a single true positive to ensure consistent evaluation. This dual scoring method accounts for differences in extraction approaches and provides a balanced assessment.






\vspace{-6pt}

\subsection{Results}

The evaluation results are reported in Table \ref{tab:totalres}. On average, the NS system achieves acceptable precision (66.6 for total, 62.3 for core), but lower recall (44.6 for total, 39.3 for core) and subpar F1 scores (52.7 for total, 47.2 for core). In contrast, the LLM-based system performs better across all metrics, with higher precision (72.5 for total, 65.8 for core), substantially better recall  (67.4 for total, 61.7 for core), and stronger F1 scores (69.4 for total, 63.0 for core). The gap is particularly pronounced in recall, where the LLM-based system significantly outperforms the NS system across all subdomains. This trend highlights the LLM-based system’s advantage in handling the variability and contextual richness of natural dialogue.

To assess the robustness of these results given the limited dataset size, we conducted statistical significance testing using bootstrap resampling (n = 10,000). We consider differences statistically significant when the 95\% confidence intervals do not substantially overlap. Based on this criterion, improvements in recall and F1 scores are statistically significant---particularly in the crop and dairy domains, where the performance gaps are most pronounced. In contrast, differences in precision are smaller and not statistically significant, as confidence intervals overlap substantially across all domains. These findings reinforce the strength of the LLM-based approach, especially in extracting information from complex, context-rich interviews where generalization and recall are critical.

\vspace{-7pt}

\section{Discussion}


While the LLM-based approach performed better, neither system fully solved the problem (see Appendix \ref{challenges} for the challenges). Error analysis showed that the NS system struggled with missing rules, identifier-value mismatches, and ontology grounding errors. The LLM-based system’s black-box nature made error tracing difficult, but common issues included topic segmentation errors and hallucinations (see Appendix \ref{error_analysis} for full discussion). Both approaches have distinct strengths and weaknesses (see Appendix \ref{sec:advantages&disadvantages} for the summary).


\subsection{Advantages and disadvantages of the neuro-symbolic system}

The NS system stands out for its exceptional efficiency and controllability, making it an attractive choice for resource-constrained or real-time applications. As shown in Table \ref{tab:runtime}, it runs dramatically faster than the LLM-based system: 69 seconds on CPU compared to over 22,000 seconds, and still faster when LLM-based system is GPU-accelerated (69 vs. 74 seconds). This performance edge enables rapid iteration and low-latency deployment—key factors in industrial settings.

Equally important is its interpretability. The NS system offers fine-grained control, allowing developers to identify and fix errors systematically via rule adjustments. This transparency is especially valuable in high-stakes domains where reliability and accountability matter. It also performs well when extraction criteria are clear and context-independent, delivering consistent accuracy.

However, these benefits come with clear limitations. Rule-based logic lacks generalizability and often fails when unseen cases fall outside the predefined rules. Our error analysis shows that the NS system consistently underperforms in tasks requiring nuanced context understanding. Moreover, rule development is labor-intensive and demands deep domain expertise, making it difficult to scale or adapt to new domains. Despite promising directions in semi-automated rule learning~\citep{noriega-atala-etal-2022-neural,vacareanu-etal-2022-human,vacareanu2022synthlrec}, manual effort remains a bottleneck.



\subsection{Advantages and disadvantages of the LLM-based system}

The LLM-based system offers greater adaptability and higher extraction quality, particularly in complex, context-rich scenarios. Its strength lies in generalization: it performs well with minimal supervision, relying only on task descriptions and a handful of examples. This makes it especially suited to evolving domains where rules are difficult to anticipate or enumerate.

\begin{table}[h!]
    \centering
    \renewcommand{\arraystretch}{0.8}
    \resizebox{\columnwidth}{!}{
    \begin{tabular}{lccc}
        \toprule
        & \textbf{Neuro-symbolic} & \textbf{LLM (GPU)} & \textbf{LLM (CPU)} \\
        \midrule
        Pork   & 66 & 66  & 14878 \\
        Crop   & 74 & 71  & 27950 \\
        Dairy  & 68 & 86  & 26133    \\
        \midrule
        Average & 69 & 74  & 22987    \\
        \bottomrule   
    \end{tabular}
    }
    \caption{Runtime comparison of neuro-symbolic and LLM-based approaches (in seconds)\protect\footnotemark}
    \label{tab:runtime}
\end{table}
\footnotetext{Experiments were conducted on an Apple M1 chip (8-core CPU, 8GB unified memory) for the CPU setting and Google Colab's NVIDIA A100 (40GB VRAM) for GPU acceleration.}

Deployment is also streamlined. Unlike the NS system, which requires laborious rule engineering, the LLM-based approach enables agile iteration via simple keyword tweaks or dataclass modifications. This lowers the barrier to entry, reduces long-term maintenance costs, and improves accessibility and scalability across domains.

Nonetheless, these advantages come at a cost. As noted earlier, the system is computationally expensive and significantly slower. More critically, it functions as a black box, offering limited transparency or control. When errors occur, diagnosing and fixing them is non-trivial. Model selection also introduces risk: if the LLM is not well-matched to the task (e.g., not optimized for dialogue), performance degrades. Finally, hallucinations remain an unresolved challenge, posing risks in applications where factual accuracy is essential.

\vspace{-5pt}

\section{Conclusion}

This study presents and compares two information extraction systems---an NS system and an LLM-based system---evaluated on agricultural interview transcripts across multiple subdomains. The result indicates that the LLM-based system outperforms the neuro-symbolic system, particularly in terms of recall, although neither approach completely resolves the task.

Overall, each system has distinct advantages and disadvantages. The NS system is efficient, offers greater control, and is highly accurate in context-free tasks. However, it lacks generalizability, struggles with contextual nuances, and demands substantial resources for development and maintenance. In contrast, the LLM-based system offers higher performance, faster deployment, and more cost-effective maintenance but is constrained by slower runtime, limited control, model dependency, and a tendency to hallucinate. We hope this discussion informs future IE system design for real-world deployment and sustainability. The dataset accompanying this study is publicly available at: \url{https://anonymous.4open.science/r/ag_dataset-8E0B/}

\section{Limitations}

While this study offers valuable insights, it has limitations. The findings are based on a single domain (agriculture) and a specific, though complex, IE task---informational interviews. Only one NS and one LLM architecture were evaluated. Although the dataset includes just nine interviews, each is lengthy and content-rich, averaging 28 minutes 27 seconds of audio, 4,700 words, and 189 speaker turns. This makes the dataset a dense and meaningful testbed despite the limited number of interviews. Future work should examine broader domains, additional IE tasks, and diverse model architectures with larger datasets to validate and extend these findings.

\section{Ethics statement}

This study uses data from real-world interviews involving human participants. To protect their privacy and confidentiality, we conducted a thorough manual anonymization process. Each interview was reviewed to ensure that no personally identifiable information is present. Additionally, sensitive numerical details (e.g., farm sizes, production volumes) were replaced with plausible but non-identifiable values to prevent reidentification while preserving the utility of the data. The resulting dataset, fully anonymized and privacy-safe, is made publicly available to support further research: \url{https://anonymous.4open.science/r/ag_dataset-8E0B/}








\bibliography{anthology,custom}

\begin{thebibliography}{31}
\expandafter\ifx\csname natexlab\endcsname\relax\def\natexlab#1{#1}\fi

\bibitem[{Appelt and Onyshkevych(1998)}]{10.3115/1119089.1119095}
Douglas~E. Appelt and Boyan Onyshkevych. 1998.
\newblock \href {https://doi.org/10.3115/1119089.1119095} {The common pattern
  specification language}.
\newblock In \emph{Proceedings of a Workshop on Held at Baltimore, Maryland:
  October 13-15, 1998}, TIPSTER '98, page 23–30, USA. Association for
  Computational Linguistics.

\bibitem[{Chen et~al.(2015)Chen, Xu, Liu, Zeng, and
  Zhao}]{chen-etal-2015-event}
Yubo Chen, Liheng Xu, Kang Liu, Daojian Zeng, and Jun Zhao. 2015.
\newblock \href {https://doi.org/10.3115/v1/P15-1017} {Event extraction via
  dynamic multi-pooling convolutional neural networks}.
\newblock In \emph{Proceedings of the 53rd Annual Meeting of the Association
  for Computational Linguistics and the 7th International Joint Conference on
  Natural Language Processing (Volume 1: Long Papers)}, pages 167--176,
  Beijing, China. Association for Computational Linguistics.

\bibitem[{Chiticariu et~al.(2013)Chiticariu, Li, and
  Reiss}]{chiticariu-etal-2013-rule}
Laura Chiticariu, Yunyao Li, and Frederick~R. Reiss. 2013.
\newblock \href {https://aclanthology.org/D13-1079/} {Rule-based information
  extraction is dead! long live rule-based information extraction systems!}
\newblock In \emph{Proceedings of the 2013 Conference on Empirical Methods in
  Natural Language Processing}, pages 827--832, Seattle, Washington, USA.
  Association for Computational Linguistics.

\bibitem[{Colvin et~al.(2025)Colvin, Jolibois, Ramezani, Garcia~Badaracco,
  Dorsey, Montague, Matveenko, Trylesinski, Runkle, Hewitt, Hall, and
  Plot}]{Colvin_Pydantic_2025}
Samuel Colvin, Eric Jolibois, Hasan Ramezani, Adrian Garcia~Badaracco, Terrence
  Dorsey, David Montague, Serge Matveenko, Marcelo Trylesinski, Sydney Runkle,
  David Hewitt, Alex Hall, and Victorien Plot. 2025.
\newblock \href {https://github.com/pydantic/pydantic} {{Pydantic}}.

\bibitem[{Cunningham et~al.(2002)Cunningham, Maynard, Bontcheva, and
  Tablan}]{cunningham-etal-2002-gate}
Hamish Cunningham, Diana Maynard, Kalina Bontcheva, and Valentin Tablan. 2002.
\newblock \href {https://doi.org/10.3115/1073083.1073112} {{GATE}: an
  architecture for development of robust {HLT} applications}.
\newblock In \emph{Proceedings of the 40th Annual Meeting of the Association
  for Computational Linguistics}, pages 168--175, Philadelphia, Pennsylvania,
  USA. Association for Computational Linguistics.

\bibitem[{Danilevsky et~al.(2021)Danilevsky, Dhanorkar, Li, Popa, Qian, and
  Xu}]{danilevsky_2021_explainability_for_nlp}
Marina Danilevsky, Shipi Dhanorkar, Yunyao Li, Lucian Popa, Kun Qian, and
  Anbang Xu. 2021.
\newblock \href {https://doi.org/10.1145/3447548.3470808} {Explainability for
  natural language processing}.
\newblock In \emph{Proceedings of the 27th ACM SIGKDD Conference on Knowledge
  Discovery \&amp; Data Mining}, KDD '21, page 4033–4034, New York, NY, USA.
  Association for Computing Machinery.

\bibitem[{Gardner et~al.(2013)Gardner, Talukdar, Kisiel, and
  Mitchell}]{gardner-etal-2013-improving}
Matt Gardner, Partha~Pratim Talukdar, Bryan Kisiel, and Tom Mitchell. 2013.
\newblock \href {https://aclanthology.org/D13-1080/} {Improving learning and
  inference in a large knowledge-base using latent syntactic cues}.
\newblock In \emph{Proceedings of the 2013 Conference on Empirical Methods in
  Natural Language Processing}, pages 833--838, Seattle, Washington, USA.
  Association for Computational Linguistics.

\bibitem[{Hobbs et~al.(1993)Hobbs, Appelt, Bear, Israel, Kameyalna, and
  Tyson}]{hobbs-etal-1993-fastus}
Jerry~R. Hobbs, Douglas Appelt, John Bear, David Israel, Megumi Kameyalna, and
  Mabry Tyson. 1993.
\newblock \href {https://aclanthology.org/H93-1026/} {{FASTUS}: A system for
  extracting information from text}.
\newblock In \emph{{H}uman {L}anguage {T}echnology: Proceedings of a Workshop
  Held at Plainsboro, New Jersey, March 21-24, 1993}.

\bibitem[{Hu et~al.(2025)Hu, Li, Jin, Bai, Guo, and Cheng}]{hu-etal-2025-large}
Zhilei Hu, Zixuan Li, Xiaolong Jin, Long Bai, Jiafeng Guo, and Xueqi Cheng.
  2025.
\newblock \href {https://aclanthology.org/2025.coling-main.500/} {Large
  language model-based event relation extraction with rationales}.
\newblock In \emph{Proceedings of the 31st International Conference on
  Computational Linguistics}, pages 7484--7496, Abu Dhabi, UAE. Association for
  Computational Linguistics.

\bibitem[{Imaichi et~al.(2013)Imaichi, Yanase, and
  Niwa}]{imaichi-etal-2013-comparison}
Osamu Imaichi, Toshihiko Yanase, and Yoshiki Niwa. 2013.
\newblock \href {https://aclanthology.org/W13-4607/} {A comparison of
  rule-based and machine learning methods for medical information extraction}.
\newblock In \emph{The First Workshop on Natural Language Processing for
  Medical and Healthcare Fields}, pages 38--42, Nagoya. Asian Federation of
  Natural Language Processing.

\bibitem[{Josifoski et~al.(2022)Josifoski, De~Cao, Peyrard, Petroni, and
  West}]{josifoski-etal-2022-genie}
Martin Josifoski, Nicola De~Cao, Maxime Peyrard, Fabio Petroni, and Robert
  West. 2022.
\newblock \href {https://doi.org/10.18653/v1/2022.naacl-main.342} {{G}en{IE}:
  Generative information extraction}.
\newblock In \emph{Proceedings of the 2022 Conference of the North American
  Chapter of the Association for Computational Linguistics: Human Language
  Technologies}, pages 4626--4643, Seattle, United States. Association for
  Computational Linguistics.

\bibitem[{Liu et~al.(2020)Liu, Gardner, Cohen, and
  Lapata}]{liu-etal-2020-multi}
Jiangming Liu, Matt Gardner, Shay~B. Cohen, and Mirella Lapata. 2020.
\newblock \href {https://doi.org/10.18653/v1/2020.emnlp-main.245} {Multi-step
  inference for reasoning over paragraphs}.
\newblock In \emph{Proceedings of the 2020 Conference on Empirical Methods in
  Natural Language Processing (EMNLP)}, pages 3040--3050, Online. Association
  for Computational Linguistics.

\bibitem[{Ma et~al.(2023)Ma, Gong, He, Zhao, and Duan}]{ma-etal-2023-query}
Xinbei Ma, Yeyun Gong, Pengcheng He, Hai Zhao, and Nan Duan. 2023.
\newblock \href {https://doi.org/10.18653/v1/2023.emnlp-main.322} {Query
  rewriting in retrieval-augmented large language models}.
\newblock In \emph{Proceedings of the 2023 Conference on Empirical Methods in
  Natural Language Processing}, pages 5303--5315, Singapore. Association for
  Computational Linguistics.

\bibitem[{Nguyen and Grishman(2015)}]{nguyen-grishman-2015-relation}
Thien~Huu Nguyen and Ralph Grishman. 2015.
\newblock \href {https://doi.org/10.3115/v1/W15-1506} {Relation extraction:
  Perspective from convolutional neural networks}.
\newblock In \emph{Proceedings of the 1st Workshop on Vector Space Modeling for
  Natural Language Processing}, pages 39--48, Denver, Colorado. Association for
  Computational Linguistics.

\bibitem[{Noriega-Atala et~al.(2022)Noriega-Atala, Vacareanu, Hahn-Powell, and
  Valenzuela-Esc{\'a}rcega}]{noriega-atala-etal-2022-neural}
Enrique Noriega-Atala, Robert Vacareanu, Gus Hahn-Powell, and Marco~A.
  Valenzuela-Esc{\'a}rcega. 2022.
\newblock \href {https://aclanthology.org/2022.pandl-1.10} {Neural-guided
  program synthesis of information extraction rules using self-supervision}.
\newblock In \emph{Proceedings of the First Workshop on Pattern-based
  Approaches to NLP in the Age of Deep Learning}, pages 85--93, Gyeongju,
  Republic of Korea. International Conference on Computational Linguistics.

\bibitem[{Perot et~al.(2024)Perot, Kang, Luisier, Su, Sun, Boppana, Wang, Wang,
  Mu, Zhang, Lee, and Hua}]{perot-etal-2024-lmdx}
Vincent Perot, Kai Kang, Florian Luisier, Guolong Su, Xiaoyu Sun, Ramya~Sree
  Boppana, Zilong Wang, Zifeng Wang, Jiaqi Mu, Hao Zhang, Chen-Yu Lee, and Nan
  Hua. 2024.
\newblock \href {https://doi.org/10.18653/v1/2024.findings-acl.899} {{LMDX}:
  Language model-based document information extraction and localization}.
\newblock In \emph{Findings of the Association for Computational Linguistics:
  ACL 2024}, pages 15140--15168, Bangkok, Thailand. Association for
  Computational Linguistics.

\bibitem[{Sculley et~al.(2015)Sculley, Holt, Golovin, Davydov, Phillips, Ebner,
  Chaudhary, Young, Crespo, and Dennison}]{Sculley2015HiddenTD}
D.~Sculley, Gary Holt, Daniel Golovin, Eugene Davydov, Todd Phillips, Dietmar
  Ebner, Vinay Chaudhary, Michael Young, Jean-François Crespo, and Dan
  Dennison. 2015.
\newblock Hidden technical debt in machine learning systems.
\newblock In \emph{NIPS}.

\bibitem[{Socher et~al.(2012)Socher, Huval, Manning, and
  Ng}]{socher-etal-2012-semantic}
Richard Socher, Brody Huval, Christopher~D. Manning, and Andrew~Y. Ng. 2012.
\newblock \href {https://aclanthology.org/D12-1110/} {Semantic compositionality
  through recursive matrix-vector spaces}.
\newblock In \emph{Proceedings of the 2012 Joint Conference on Empirical
  Methods in Natural Language Processing and Computational Natural Language
  Learning}, pages 1201--1211, Jeju Island, Korea. Association for
  Computational Linguistics.

\bibitem[{Stutz et~al.(2015)Stutz, Paudel, Verman, and
  Bernstein}]{10.1145/2736277.2741687}
Philip Stutz, Bibek Paudel, Mihaela Verman, and Abraham Bernstein. 2015.
\newblock \href {https://doi.org/10.1145/2736277.2741687} {Random walk
  triplerush: Asynchronous graph querying and sampling}.
\newblock In \emph{Proceedings of the 24th International Conference on World
  Wide Web}, WWW '15, page 1034–1044, Republic and Canton of Geneva, CHE.
  International World Wide Web Conferences Steering Committee.

\bibitem[{Sundheim(1991)}]{sundheim1991overview}
Beth~M Sundheim. 1991.
\newblock Overview of the third message understanding evaluation and
  conference.
\newblock In \emph{Third Message Understanding Conference (MUC-3): Proceedings
  of a Conference Held in San Diego, California, May 21-23, 1991}.

\bibitem[{Tang and Surdeanu(2023)}]{Tang2022ItTT}
Zheng Tang and Mihai Surdeanu. 2023.
\newblock \href {https://arxiv.org/pdf/2204.11424.pdf} {It takes two flints to
  make a fire: Multitask learning of neural relation and explanation
  classifiers}.
\newblock \emph{Computational Linguistics}.
\newblock Accepted on 2022.

\bibitem[{Vacareanu et~al.(2024)Vacareanu, Alam, Islam, Riaz, and
  Surdeanu}]{vacareanu2024softrules}
Robert Vacareanu, Fahmida Alam, Md~Asiful Islam, Haris Riaz, and Mihai
  Surdeanu. 2024.
\newblock \href {https://arxiv.org/pdf/2403.03305.pdf} {Best of both worlds: A
  pliable and generalizable neuro-symbolic approach for relation
  classification}.
\newblock In \emph{Findings of the Association for Computational Linguistics:
  NAACL 2024}, Mexico City, Mexico. Association for Computational Linguistics.

\bibitem[{Vacareanu et~al.(2022{\natexlab{a}})Vacareanu, Barbosa,
  Noriega-Atala, Hahn-Powell, Sharp, Valenzuela-Esc{\'a}rcega, and
  Surdeanu}]{vacareanu-etal-2022-human}
Robert Vacareanu, George~C.G. Barbosa, Enrique Noriega-Atala, Gus Hahn-Powell,
  Rebecca Sharp, Marco~A. Valenzuela-Esc{\'a}rcega, and Mihai Surdeanu.
  2022{\natexlab{a}}.
\newblock \href {https://aclanthology.org/2022.naacl-demo.8} {A human-machine
  interface for few-shot rule synthesis for information extraction}.
\newblock In \emph{Proceedings of the 2022 Conference of the North American
  Chapter of the Association for Computational Linguistics: Human Language
  Technologies: System Demonstrations}, pages 64--70, Hybrid: Seattle,
  Washington + Online. Association for Computational Linguistics.

\bibitem[{Vacareanu et~al.(2020)Vacareanu, Gouveia~Barbosa,
  Valenzuela-Esc{\'a}rcega, and Surdeanu}]{vacareanu-etal-2020-parsing}
Robert Vacareanu, George~Caique Gouveia~Barbosa, Marco~A.
  Valenzuela-Esc{\'a}rcega, and Mihai Surdeanu. 2020.
\newblock \href {https://aclanthology.org/2020.lrec-1.643/} {Parsing as
  tagging}.
\newblock In \emph{Proceedings of the Twelfth Language Resources and Evaluation
  Conference}, pages 5225--5231, Marseille, France. European Language Resources
  Association.

\bibitem[{Vacareanu et~al.(2022{\natexlab{b}})Vacareanu,
  Valenzuela-Esc\'{a}rcega, Barbosa, Sharp, Hahn-Powell, and
  Surdeanu}]{vacareanu2022synthlrec}
Robert Vacareanu, Marco~A. Valenzuela-Esc\'{a}rcega, George Barbosa, Rebecca
  Sharp, Gus Hahn-Powell, and Mihai Surdeanu. 2022{\natexlab{b}}.
\newblock \href
  {http://www.lrec-conf.org/proceedings/lrec2022/pdf/2022.lrec-1.665.pdf} {From
  examples to rules: Neural guided rule synthesis for information extraction}.
\newblock In \emph{Proceedings of the 13th Language Resources and Evaluation
  Conference (LREC)}.

\bibitem[{Valenzuela-Esc{\'a}rcega et~al.(2015)Valenzuela-Esc{\'a}rcega,
  Hahn-Powell, Surdeanu, and Hicks}]{valenzuela-escarcega-etal-2015-domain}
Marco~A. Valenzuela-Esc{\'a}rcega, Gus Hahn-Powell, Mihai Surdeanu, and Thomas
  Hicks. 2015.
\newblock \href {https://doi.org/10.3115/v1/P15-4022} {A domain-independent
  rule-based framework for event extraction}.
\newblock In \emph{Proceedings of {ACL}-{IJCNLP} 2015 System Demonstrations},
  pages 127--132, Beijing, China. Association for Computational Linguistics and
  The Asian Federation of Natural Language Processing.

\bibitem[{Wadhwa et~al.(2023)Wadhwa, Amir, and
  Wallace}]{wadhwa-etal-2023-revisiting}
Somin Wadhwa, Silvio Amir, and Byron Wallace. 2023.
\newblock \href {https://doi.org/10.18653/v1/2023.acl-long.868} {Revisiting
  relation extraction in the era of large language models}.
\newblock In \emph{Proceedings of the 61st Annual Meeting of the Association
  for Computational Linguistics (Volume 1: Long Papers)}, pages 15566--15589,
  Toronto, Canada. Association for Computational Linguistics.

\bibitem[{Wang et~al.(2024)Wang, Huang, Xu, and Lu}]{wang2024does}
Xin Wang, Liangliang Huang, Shuozhi Xu, and Kun Lu. 2024.
\newblock How does a generative large language model perform on domain-specific
  information extraction?- a comparison between gpt-4 and a rule-based method
  on band gap extraction.
\newblock \emph{Journal of Chemical Information and Modeling},
  64(20):7895--7904.

\bibitem[{Xu et~al.(2024)Xu, Chen, Peng, Zhang, Xu, Zhao, Wu, Zheng, Wang, and
  Chen}]{xu2024large}
Derong Xu, Wei Chen, Wenjun Peng, Chao Zhang, Tong Xu, Xiangyu Zhao, Xian Wu,
  Yefeng Zheng, Yang Wang, and Enhong Chen. 2024.
\newblock Large language models for generative information extraction: A
  survey.
\newblock \emph{Frontiers of Computer Science}, 18(6):186357.

\bibitem[{Zeng et~al.(2015)Zeng, Liu, Chen, and Zhao}]{zeng-etal-2015-distant}
Daojian Zeng, Kang Liu, Yubo Chen, and Jun Zhao. 2015.
\newblock \href {https://doi.org/10.18653/v1/D15-1203} {Distant supervision for
  relation extraction via piecewise convolutional neural networks}.
\newblock In \emph{Proceedings of the 2015 Conference on Empirical Methods in
  Natural Language Processing}, pages 1753--1762, Lisbon, Portugal. Association
  for Computational Linguistics.

\bibitem[{Zeng et~al.(2014)Zeng, Liu, Lai, Zhou, and
  Zhao}]{zeng-etal-2014-relation}
Daojian Zeng, Kang Liu, Siwei Lai, Guangyou Zhou, and Jun Zhao. 2014.
\newblock \href {https://aclanthology.org/C14-1220/} {Relation classification
  via convolutional deep neural network}.
\newblock In \emph{Proceedings of {COLING} 2014, the 25th International
  Conference on Computational Linguistics: Technical Papers}, pages 2335--2344,
  Dublin, Ireland. Dublin City University and Association for Computational
  Linguistics.

\end{thebibliography}
\bibliographystyle{acl_natbib}
\clearpage
\appendix

\section{Related Works}
\label{related_works}

Information Extraction (IE) is a key task in natural language processing that converts unstructured text into structured data. Early IE systems were primarily rule-based, including FASTUS \citep{hobbs-etal-1993-fastus}, CPSL \citep{10.3115/1119089.1119095}, and GATE \citep{cunningham-etal-2002-gate}. As data-driven methods advanced, neural approaches such as those by \citet{socher-etal-2012-semantic}, \citet{zeng-etal-2014-relation}, \citet{chen-etal-2015-event}, \citet{nguyen-grishman-2015-relation}, and \citet{zeng-etal-2015-distant} gained traction and became widely adopted. More recently, large language models (LLMs) have been explored for IE tasks (\citealp{josifoski-etal-2022-genie}; \citealp{ma-etal-2023-query}; \citealp{wadhwa-etal-2023-revisiting}; \citealp{wadhwa-etal-2023-revisiting}; \citealt{perot-etal-2024-lmdx}; \citealp{hu-etal-2025-large}). These models often achieve strong performance but still face challenges such as hallucination.

Alongside these developments, several works have explored combining or comparing different approaches. For instance, \citet{gardner-etal-2013-improving} used the Path Ranking Algorithm over knowledge base graphs for relation extraction, and \citet{10.1145/2736277.2741687} introduced a framework combining SPARQL querying with graph sampling techniques. \citet{liu-etal-2020-multi} proposed a compositional model that integrates symbolic structures with neural reasoning to enable complex multi-hop inference. Comparative studies include \citet{imaichi-etal-2013-comparison}, who evaluated rule-based and machine learning approaches in medical IE, finding that each method had strengths depending on task structure. \citet{wang2024does} compared GPT-4 to a rule-based system for materials science IE, showing that while GPT-4 achieved higher accuracy, it struggled with consistency and hallucinated facts.

\section{An entity rule for neuro-symbolic system}
\label{entityrule}

The figure below shows a simplified sample rule for extracting compound entities containing common identifier key words (triggers). The rule takes advantage of syntactic patterns that connect generic triggers to other nouns or noun phrases to form compound identifiers. For example, an identifier can be made of the trigger word `capacity' and a noun phrase `those barns', connected to it with a nominal modifier syntactic dependency relation.

\begin{figure}[h!]
\small
    \centering
    \setlength{\baselineskip}{0.8\baselineskip} 
    \begin{verbatim}
- name: all-generic-entity-dep
  label: GenericEntity
  example: "the capacity of those barns"
  type: "dependency"
  pattern: |
    trigger = [word=/^(capacity|number|
               usage|production|price|
               cost|consumption|date)$/]
    variable: NounPhrase =  nmod_of 
    \end{verbatim}
    \caption{A simplified sample rule for extracting compound entities that contain common identifier key words (triggers). }

\end{figure}

\section{Prompt used for the LLM-based system}
\label{sec:prompt}

Below is the prompt used for the LLM-based system. It is populated with information from the dataclass, where placeholders enclosed in curly braces (e.g., \{schema\}, \{example\}, \{text\}) indicate where specific details should be inserted.

\begin{quote}
You are an advanced AI designed to extract key agricultural information from interviews. Accuracy is crucial as it impacts crop yield, finances, and resource management. Ensure high precision and high recall in your extractions. \\

\%\%Instructions: \\
1. Read Carefully: Identify and extract only the information relevant to agriculture from the text. Remember that the text is an interview. Focus on answers than on questions when extracting information. \\
2. Exclusivity and Precision: Only include explicitly mentioned information. Avoid assumptions, and do not alter numerical data. Ensure accuracy and completeness. \\
3. Extract Exhaustively: Extract all instances of the same type of information if multiple are present. There can be two or more instances within the same text. In such cases, you should extract ALL of them (e.g., If there are three instances, return [\{\{first instance\}\}, \{\{second instance\}\}, \{\{third instance\}\}]). \\
4. No Hallucination: If no relevant information is present in the text, do not return anything. Do NOT copy/extract anything from the schema/format examples given in three backticks (```). \\
\end{quote}
\clearpage

\begin{quote}
\%\%Format:\\
1. Schema Compliance: Use the provided schema for extraction. Do not add or alter properties/fields. \\
```\{schema\}``` (do NOT extract anything from within three backticks) \\
2. JSON Format: Return the data in JSON format as shown in the example. Follow the format exactly, but do not copy its content in any case. \\
```\{example\}``` (do NOT extract anything from within three backticks) \\

\%\%Text for Extraction: \{text\} \\

Your task is essential for informed agricultural decision-making. Continuously refine your extraction techniques to stay accurate and relevant.
\end{quote}

\section{Sample dataclass}
\label{sec:dataclass}

Below is the dataclass for the total finishing pigs information. The schema and example in the dataclass is used to supplement the prompt. \\

\begin{lstlisting}
    class TotalFinishingPigsEvent(Notation):
    """
    Information about the total finishing pigs in the farm.

    Args:
        total_finishing_pigs: total number of finishing pigs in the farm. Integer type outputs expected.

    Example:
        From a statement like "And so you're finishing how many total pigs a year? 6,670.", the extracted information should be:
        [{"total_finishing_pigs": 6670}]
    """

    total_finishing_pigs: typing.Optional[int] = None
\end{lstlisting}

\section{Word Count Statistics per Interview}\label{word-count-statistics}

Table \ref{tab:wordcounts} presents word counts for nine interviews.

\vspace*{\fill} 

\begin{table}[h]
\centering
\begin{tabular}{cc}
\toprule
\textbf{Interview Title} & \textbf{Word Counts} \\
\midrule
Pork-interview-1 & 5015 \\
Pork-interview-2 & 4028 \\
Pork-interview-3 & 4650 \\
Crop-interview-1 & 4332 \\
Crop-interview-2 & 4006 \\
Crop-interview-3 & 6839 \\
Dairy-interview-1 & 1506 \\
Dairy-interview-2 & 3984 \\
Dairy-interview-3 & 7939 \\
\midrule
\textbf{Average} & \textbf{4700} \\
\bottomrule
\end{tabular}
\caption{Interview Word Count Statistics}
\label{tab:wordcounts}
\end{table}

\section{Challenges}
\label{challenges}

The nature of the data that we work with creates a number of challenges for an information extraction system. We subdivide these challenges into two main categories. The first category is genre-related challenges: we are working with transcripts of interviews, which means the text that we are extracting information from is imperfect both in how it is an automatically recognized transcription of the original, and in how it is transcription of practically unscripted interviews, with various pragmatic, conversation-discourse-related, and speaker-related issues arising from that. The second category is domain-related challenges: these issues have to do with the specific domain that we are working with---agriculture---and specific needs of the project. These two categories of challenges are described in more detail in the next sections. 

\subsection{Genre-related issues}

Genre-related issues have to do with the fact that the text we work with is both an automatic transcript and a semi-scripted interview. The issues described below should apply to any text of this type regardless of the domain and specific project. 

\subsubsection{Transcript-related issues.}

The transcripts that we extract text from are produced by an ASR system. This results in errors of several types, all of which impact the IE system.

\paragraph{\textit{Word-level errors.}} The ASR system has issues with domain-specific vocabulary, site names, and spoken numbers. As an example of domain-specific terminology issue, `sow'---a term that is used to describe a pig that has had a litter and that is frequently used in the pork industry---is rarely recognized correctly by the ASR and is instead transcribed as similar sounding words, e.g., `SOS', `sal', and `South'. Site names cause issues since they are frequently proper names or have alphanumerical IDs, with parts of the ID transcribed as separate tokens because of being spelled out. The latter is also true for other numerical information: complex numbers get broken up during transcription, e.g., in the transcripts, we see `two 20' instead of `220', `2 40' instead of `240', `20, 20` instead of the year `2020',  `12 hundred' instead of `1200', and `between 13 and 1500 pounds' in the sense of `1300-1500'. These types of transcription errors require complicated normalization, which can become particularly problematic in potential cases of ambiguity, e.g., ``We have two 20-head barns'' vs. ``We have 220-head barns.''


\paragraph{\textit{Sentence-level errors.} } The ASR system does not always reliably detect questions and breaks up sentences between speaker turns, possibly because of speech overlaps, pauses, speaker intonation, recording quality, and other reasons. In the following example, the utterance attributed to Speaker 2 is a continuation of the previous speaker's utterance: 

\begin{enumerate}[leftmargin=*]
    \item \label{split}

\begin{itemize}[leftmargin=*]
    \item[] Speaker 1	00:20:19	[\dots] And did you have any interns on the farm
\item[] Speaker 2	00:20:39	This past year?  
\item[] Speaker 1	00:20:41	Uh, yeah.  

\end{itemize}
\end{enumerate}

\paragraph{\textit{Speaker diarization.} } The ASR system cannot always correctly detect which utterance belongs to which speaker. Together with incorrect segmentation of utterances (e.g., splitting an utterance between speaker turns as in Example \ref{split} of this section), this means that one speaker's utterances are not labeled consistently. Improving speaker diarization could help the system is two major ways. First, it would allow for locating identifiers with higher confidence since crucial identifiers are mentioned in the interviewer questions. Second, it would let us eliminate identifier-value pairs that are brought up by interviewers as sample answers to their questions and should not be part of the output. 

\subsubsection{Interview-related issues.} 

Some of the issues we encounter are due to the genre that we work with: spoken, semi-structured interviews. 

\paragraph{\textit{Separation of target information.}} The main issue that impacts the information extraction system, particularly because it is the extraction of relations, is the fact that in conversations, the target entities are not necessarily adjacent. They can occur within the same sentence, which could be easily handled by our rule-based extraction component, but they can also be separated by several sentences or even several conversation turns, especially when there are more than two participants. In Figure \ref{fig:canon-example}, the interviewee provides additional information about the target identifier three speaker turns after it is first mentioned.


\paragraph{\textit{Speaker-related issues. }} Several issues that we observe arise from the fact that we work with spoken language data. For instance, various speech disfluencies and errors impact the ASR output quality, e.g., the speaker repeating numbers multiple times (``We have [\dots] 4 4300 finishing'') creates a potentially ambiguous target value since it is unclear whether the first `4' is meaningful, extractable content or it is a speech production error and needs to be removed. 

Regional and individual pronunciation differences can impact the ASR system, but they can also create additional content to extract from by requiring the interview participants to clarify misunderstandings, potentially leading to more errors. In one such case, for instance, the participants had to clarify a misunderstanding that resulted from the difference in pronunciation of the term `culled'.


Finally, participating in  interviews is quite demanding, with participants needing to recall or look up a lot of information. In some cases, that results in either going back and forth on some information (Example \ref{eightorten}) or thinking out loud (Example \ref{counting}). 

\begin{enumerate}
\setcounter{enumi}{1}
    \item \label{eightorten} Uh, uh, I think it's seven. Seven or 9. Um, let's go with seven. 
    \item  \label{counting} We've got we have one 2, 3, 4, we got five [demographic] that work for us
\end{enumerate}

These types of interviewee responses are particularly difficult for NS systems, since there are multiple responses to pick from for the same question and the need to filter out all but one.




\subsection{Domain-specific issues}

The issues discussed in this section are mainly relevant to those working in specialized domains. The most prominent issue is the presence of specialized vocabulary. As discussed above, it can impact the ASR performance and also requires maintenance of lexicons and knowledge bases if we deal with a rule-based extraction system. 

Two additional issues were primarily project-specific. Certain identifiers are very semantically similar (e.g., `water used'---the amount of water used,---and `water source'). Cases like this make it harder to correctly map identifiers to the correct nodes in the ontology. Finally, in some interviews for this project, locations were discussed in relation to other locations, with interview participants using a map as scaffolding for the discussion. Our text-based system cannot handle such multi-modal cases. 





\subsection{Solutions}

While working on the project, we approached solving the issues discussed above in two ways. On the one hand, we were looking at technical solutions, for instance, we mapped frequently mis-transcribed words to the proper terms to improve the quality of the ASR output, segmented transcripts by topic, applied various filtering, and implemented the LLM-based version of the pipeline. 

At the same time, some of the issues could only be addressed on the interview participants' end. Together with the domain experts, we devised a set of recommendations for the interviewers to improve  interview consistency even under the conditions of a relatively free-flowing conversation. The recommendations included asking one question at a time, reiterating the interviewee's answer, and refocusing the discussion on the topic of interest by using such phrases as `circling back' and `jumping back in' to help with topic segmentation. Additionally, we suggested that it may be advisable to ask interviewees to prepare their records in advance to make sure they can easily look up the answers. 


\section{Error analysis}\label{error_analysis}

Error analysis showed that the NS system struggled with missing rules, identifier-value mismatches, and ontology grounding errors. The LLM-based system’s black-box nature made error tracing difficult, but common issues included topic segmentation errors and hallucinations.

\subsection{Errors of the neuro-symbolic system}

\paragraph{Absence of applicable rules:} The neuro-symbolic system relies on predefined rules and a knowledge base (KB), which limits its flexibility in handling conversational variability. In our interview data, where speakers often use informal or unexpected phrasing, the system frequently fails to extract information unless it matches a known pattern. For instance, in one crop interview, it missed herbicide names not present in the KB. Even when the correct information is mentioned, the system fails to extract it if it's phrased differently than expected---highlighting a core limitation of rule-based methods in open-ended, real-world dialogue.

\vspace{-6pt}

\paragraph{Mismatches between identifiers and values:} Despite heuristics for matching values to their correct identifiers, the system sometimes fails due to limited contextual awareness. A particularly challenging case occurs when values lack units. To prevent false positives, unit-less values are not matched by default, except for indicators that commonly appear without units (e.g., employees: "How many employees do you have?" "Three."). Issues arise when indicators that typically require units are given without them. For instance, in a dairy interview, protein and butterfat percents were stated as 4 and 3 without units. Although context suggests 4\% and 3\%, the system failed to match them due to the unit requirement.

\vspace{-6pt}

\paragraph{Incorrect ontology grounding:} In domain-specific settings like ours, semantic categories often overlap, making precise grounding especially challenging. Our system uses vector embedding similarity and string overlap to match extracted indicators and values to ontology entries. However, when multiple categories are closely related, this approach can backfire. For example, in a dairy interview, the pair `manure' (indicator) and `pits' (value) was incorrectly grounded to \textit{Manure Handling Techniques} instead of the correct \textit{Manure Storage}. Both categories shared high semantic similarity, but the wrong match scored slightly higher based on the system’s heuristics---highlighting the difficulty of disambiguation in tightly clustered ontologies.

\subsection{Errors of the LLM-based system}

\paragraph{Topic segmentation errors:}


A key challenge in applying LLMs to our data is the large context size of each interview, which exceeds typical input limits and requires prior segmentation. To manage this, we use a keyword-guided approach to help the LLM focus on relevant information within each segment (see Section \ref{sec:fine_segmentation} for details). This reduces hallucinations by narrowing the expected information scope and allows users to refine extractions by adjusting keywords.

However, this strategy introduces its own risks. Missing or misleading keywords can yield incorrect results. In one interview, certain crop names were undetected due to absent keywords. In another instance, \textit{radish} was incorrectly extracted as a tillage type because \textit{tillage radish} (a cover crop name) contained the keyword \textit{tillage}. These errors highlight a key trade-off: our reliance on keywords, rather than a statistical classifier, gives us greater control but comes at the cost of precision, as decisions are not contextually informed.

\paragraph{Hallucinations:}

The LLM-based approach is prone to hallucinations. Although we have a hallucination filter in place, it does not catch all instances. It only applies to string-type fields, allowing hallucinations in boolean or integer values to go undetected. Even for strings, the filter removes only values with no overlapping words in the source interview text to account for minor variations introduced by the generative model. However, under the current heuristics, it fails to exclude hallucinated values if they share even a single word with the source. For example, in a crop interview, hallucinated crop rotation information was not filtered out because it included \textit{years}, a word also present in the interview block.

\section{Advantages and disadvantages of neuro-symbolic and LLM-based approaches}
\label{sec:advantages&disadvantages}

Table \ref{tab:advantages&disadvantages} summarizes advantages and disadvantages of the two systems. 

\begin{table*}[hbtp!]
    \small
    \renewcommand{\arraystretch}{0.8} 
    \setlength{\tabcolsep}{5pt} 
    \begin{tabular}{m{2.5cm} p{6cm} p{6cm}} 
        \toprule
        & \hspace{10pt} \textbf{Neuro-symbolic} & \hspace{10pt} \textbf{LLM-based} \\ 
        \midrule
        \addlinespace[5pt] 
        \bfseries Advantages & 
        \begin{minipage}[c]{\linewidth}
        \begin{itemize}\setlength{\itemsep}{2pt} \setlength{\parskip}{2pt}
            \item Faster runtime
            \item Greater control
            \item High accuracy in context-free tasks
        \end{itemize} 
        \end{minipage} 
        & 
        \begin{minipage}[c]{\linewidth}
        \begin{itemize}\setlength{\itemsep}{2pt} \setlength{\parskip}{2pt}
            \item Higher performance
            \item Faster deployment
            \item Easier and cost-effective maintenance
        \end{itemize}
        \end{minipage} \\
        \addlinespace[5pt] 
        \midrule
        \addlinespace[5pt] 
        \bfseries Disadvantages & 
        \begin{minipage}[c]{\linewidth}
        \begin{itemize}\setlength{\itemsep}{2pt} \setlength{\parskip}{2pt}
            \item Limited generalizability
            \item Struggles with contextual nuances
            \item Requires significant resources to develop and maintain
        \end{itemize}
        \end{minipage} 
        & 
        \begin{minipage}[c]{\linewidth}
        \begin{itemize}\setlength{\itemsep}{2pt} \setlength{\parskip}{2pt}
            \item Slower runtime
            \item Limited control
            \item Model dependency
            \item Risk of hallucination
        \end{itemize}
        \end{minipage} \\
        \addlinespace[5pt] 
        \bottomrule
    \end{tabular}
    \caption{Comparison of the neuro-symbolic and the LLM-based approaches in terms of their advantages and disadvantages.}
    \label{tab:advantages&disadvantages}
\end{table*}

\end{document}